\documentclass[journal]{IEEEtran}

\usepackage{amsmath,amsthm}
\usepackage{subfigure}
\usepackage{amssymb}
\usepackage{amsfonts}
\usepackage{amssymb}
\usepackage{amsmath}
\usepackage{booktabs}
\usepackage{breqn}
\usepackage[numbers]{natbib}

\usepackage{bm}
\usepackage{xcolor}
\usepackage{graphicx}

\usepackage[ruled,vlined]{algorithm2e}
\usepackage{algorithmic}

\usepackage{multirow}
\usepackage{lscape}
\usepackage{epstopdf}
\usepackage{lineno}
\usepackage{subfigure}
\usepackage{multirow}
\usepackage{makecell}
\usepackage{microtype}
\usepackage{lineno}
\usepackage{url}

\ifCLASSINFOpdf
\else
\fi

\begin{document}

\title{Spatial Aggregation and Temporal Convolution Networks for Real-time Kriging}

\author{Yuankai Wu, Dingyi Zhuang, Mengying Lei, Aurelie Labbe, Lijun Sun*

\thanks{This research is supported in part by the Natural Sciences and Engineering Research Council (NSERC) of Canada and in part by the Canada Foundation for Innovation (CFI). Yuankai Wu and Mengying Lei would like to thank the Institute for Data Valorisation (IVADO) for providing scholarship to support this study.}
\thanks{Y. Wu, D. Zhuang, M, Lei, and L. Sun are with the Department of Civil Engineering, McGill University, Montreal, QC, H3A 0C3, Canada. }
\thanks{A. Labbe is with the Department of Decision Science, HEC Montreal, Montreal, QC, H3T 2A7, Canada. }
\thanks{Corresponding author: Lijun Sun (Email: lijun.sun@mcgill.ca).}
}

\maketitle

\begin{abstract}
Spatiotemporal kriging is an important application in spatiotemporal data analysis, aiming to recover/interpolate signals for unsampled/unobserved locations based on observed signals. The principle challenge for spatiotemporal kriging is how to effectively model and leverage the spatiotemporal dependencies within the data. Recently, graph neural networks (GNNs) have shown great promise for spatiotemporal kriging tasks. However, standard GNNs often require a carefully designed adjacency matrix and specific aggregation functions, which are inflexible for general applications/problems. To address this issue, we present SATCN---Spatial Aggregation and Temporal Convolution Networks---a universal and flexible framework to perform spatiotemporal kriging for various spatiotemporal datasets without the need for model specification. Specifically, we propose a novel spatial aggregation network (SAN) inspired by Principal Neighborhood Aggregation, which uses multiple aggregation functions to help one node gather diverse information from its neighbors. To exclude information from unsampled nodes, a masking strategy that prevents the unsampled sensors from sending messages to their neighborhood is introduced to SAN. We capture temporal dependencies by the temporal convolutional networks, which allows our model to cope with data of diverse sizes. To make SATCN generalizable to unseen nodes and even unseen graph structures, we employ an inductive strategy to train SATCN. We conduct extensive experiments on three real-world spatiotemporal datasets, including traffic speed and climate recordings. Our results demonstrate the superiority of SATCN over traditional and GNN-based kriging models.
\end{abstract}


\IEEEpeerreviewmaketitle

\section{Introduction}

Spatiotemporal data is ubiquitous in many domains, including traffic estimation/prediction, climate modeling, neuroscience, and earth sciences \citep{cressie2015statistics,banerjee2014hierarchical,atluri2018spatio}. In general, spatiotemporal data characterizes how a particular variable or a group of variables vary in both space and time. For example, sea surface temperature data can help climate researchers understand the effect of climate change and identify abnormal weather patterns; traffic speed data collected from a network of sensors can reveal the evolution of traffic congestion on a highway network over time. A unique property of spatiotemporal data is that the variables often show strong dependencies within/across both the spatial and the temporal dimensions. This property makes the data instances structurally correlated with each other. How to effectively model these dependencies and relations is a critical challenge in spatiotemporal data analysis.

This paper focuses on the \textit{spatiotemporal kriging} problem, of which the goal is to perform signal interpolation for unsampled locations given the signals from sampled locations during the same period. The most prevalent method for kriging is Gaussian Process (GP) regression \citep{williams2006gaussian}, which captures the correlation among all data points using a carefully specified kernel/covariance structure. For spatiotemporal problems, one can design an appropriate kernel structure to capture the correlations both within and across the spatial and temporal dimensions (see e.g., \citep{luttinen2012efficient}). However, GP is computationally very expensive: the time complexity of GP models scales cubically with the number of data points, and learning accurate hyperparameters is very challenging for complex kernel structures with computational issues such as local optima. An alternative to GP on large-scale dataset is low-rank matrix/tensor completion  \citep{bahadori2014fast,yu2016temporal,takeuchi2017autoregressive}. These models essentially impose a low-rank assumption to capture the global consistency in the data, and further include regularization structures to also ensure local consistency (e.g., autoregressive regularizer for temporal smoothness \citep{yu2016temporal} and Laplacian regularizer for spatial smoothness \citep{bahadori2014fast}). These models provide a powerful solution for missing data imputation when data is missing at random.  However, matrix/tensor completion is essentially transductive \citep{zhang2019inductive}: for a new spatial location, we have to retrain the full model even in response to only minor changes in the spatial and temporal dimension \citep{wu2020inductive}. In addition, spatiotemporal kriging corresponds to a challenging whole-row/column missing scenario in a spatiotemporal matrix, and thus model accuracy relies heavily on the specification of the local spatial Laplacian regularizer.

Recently, deep learning models have opened new doors to spatiotemporal data analysis. In general, spatiotemporal data generated from a sensor network can be naturally modeled as time-varying signals on a spatial graph structure. Numerous neural network architectures for learning on graphs \citep{KipfW17, defferrard2016convolutional, xu2018powerful} have been proposed in recent years, and graph neural networks (GNN) have been widely applied to modeling spatiotemporal data. Although the spatiotemporal kriging problem can be considered a ``forecasting'' problem along the spatial dimension, most existing GNN-based studies only focus on the multivariate time series forecasting problem on a fixed spatial graph \citep{yu2018spatio, li2018diffusion, wu2019graph, dai2020hybrid}, which cannot be generalized to model unseen locations in spatially varying graphs. Two recent studies have developed GNN-based models for spatiotemporal kriging: Kriging Convolutional Networks (KCN) \citep{appleby2020kriging} and Inductive Graph Neural Networks for Kriging (IGNNK) \citep{wu2020inductive}. Both studies suggest that GNNs are promising tools for the real-time spatiotemporal kriging problem; however, two challenging issues remain when applying the models to diverse real-world applications. The first issue is that all GNNs use a predefined rule to transform spatial information into an adjacency matrix (e.g., using Gaussian kernel as in \citep{wu2020inductive, appleby2020kriging}). We would argue that defining adjacency matrix in GNNs is equally important as defining a spatial kernel in GP, as the predefined rules for constructing the adjacency matrix determines how GNNs transform and aggregate information. Yet, the complex spatial dependencies make it difficult to specify an aggregation function of GNNs that can capture sufficient information of the target datasets \citep{corso2020principal}. To achieve high accuracy, the models require extensive fine-tuning of hyperparameters the control the predefined rule and the type of aggregation function. Second, both existing GNN-based kriging models do not fully utilize temporal dependencies. For example, KCN \citep{appleby2020kriging} treats observation time as an additional feature for GNNs, and thus neglects the temporal dependencies in spatiotemporal datasets. IGNNK, on the other hand, considers observations over a particular time period as features \citep{wu2020inductive}; as a result, it cannot handle inputs with different sizes of temporal windows.

To address the aforementioned limitations, we propose a general framework, Spatial Aggregation and Temporal Convolution Networks (SATCN), for spatiotemporal kriging. We utilize temporal convolutional networks (TCN) to model the temporal dependencies and make our framework flexible on both spatial and temporal dimensions. To address the tuning issue in modeling spatial dependencies, we propose a novel Spatial Aggregation Network (SAN) structure inspired by Principal Neighborhood Aggregation (PNA)---a recent aggregation framework proposed by \citep{corso2020principal}. Instead of performing aggregation based on a predefined adjacency matrix, each node in SAN aggregates the information of its $\mathcal{K}$-nearest neighbors together with corresponding distance information. In addition, SAN  allows for multiple different aggregators in a single layer. To provide SAN with generalization power for kriging tasks, we prevent those missing/unobserved nodes from supplying information in aggregation by masking. Finally, we train SATCN with the objective to reconstruct the full signals from all nodes. Our experiments on large-scale spatiotemporal datasets show that SATCN outperforms its deep learning and other conventional counterparts, suggesting that the proposed SATCN framework can better characterize spatiotemporal dependencies in diverse types of data. To summarize, the primary contributions of the paper are as follows:
\begin{itemize}
    \item We design a node masking strategy for real-time kriging tasks. This universal masking strategy naturally adapts to all GNNs for spatiotemporal modeling with missing/corrupted data.
    \item We leverage the temporal dependencies by temporal convolutional networks (TCN). With an inductive training strategy, our model can cope with data of diverse sizes on spatial and temporal dimensions.
    \item  We propose a spatial aggregation network (SAN)---an architecture combining multiple message aggregators with degree-scalers---to capture the complex spatial dependencies in large datasets.
\end{itemize}

The remainder of this paper is organized as follows: Section \ref{R:2} gives a brief review of the related work, followed by Section \ref{M:3} presenting the methodology. Then, Section \ref{E:4} details the empirical studies involving three different real-world datasets. Lastly, Section \ref{C:5} concludes this paper and offers some directions of future work.

\section{Related Work}
\label{R:2}
\subsection{Graph Neural Networks}

Graph neural networks (GNNs) are proposed to aggregate information from graph structure. Based on the information gathering mechanism, GNNs can be categorized into spectral approaches and spatial approaches. The essential operator in spectral GNN is the graph Laplacian, which defines graph convolutions as linear operators that diagonalize in the graph Laplacian operator \citep{mallat1999wavelet}. The generalized spectral GNN was first introduced in \cite{bruna2014spectral}. Then, \citet{defferrard2016convolutional} proposed to use Chebyshev polynomial filters on the eigenvalues to approximate the convolutional filters. Most of the state-of-the-art deep learning models for spatiotemporal data \citep{yu2018spatio, li2018diffusion,wu2019graph} are based on the concept of Chebynet. In \citep{wu2020inductive}, the authors intended to train a spectral GNN for inductive kriging task, indicating the effect of GNN for modeling spatial dependencies. However, in all spectral based approaches, the learned networks are dependent on the Laplacian matrix. This brings two drawbacks to spectral based approaches: 1) They are computationally expensive as the information aggregation has to be made in the whole graph. 2) a GNN trained on a specific structure could not be directly generalized to a graph with a different structure.

In contrast, spatial GNN approaches directly perform information aggregation on spatially close neighbors. In general, the commonalities between representative spatial GNNs can be abstracted as the following message passing mechanism \citep{gilmer2017neural}:
\begin{equation}
    \begin{split}
        a^l_{v} &= \operatorname{AGGREGATE}^{\left(l\right)}\left(\{x^l_u: u \in N(v) \}\right), \\
        x^l_{v} &= \operatorname{COMBINE}^{\left(l\right)}\left(x^{l-1}_{v},a^l_{v}\right),
    \end{split}
\end{equation}
where $x^l_v$ is the feature vector of node $v$ at the $l$-th layer, $N(v)$ is a set of nodes adjacent to $v$, and  $\operatorname{AGGREGATE}$ and $\operatorname{COMBINE}$ are parameterized functions. Here, $x^l_u$ is the message of node $u$ passing to its neighbors. Each node aggregates messages from their neighboring nodes to compute the next message. Spatial approaches have produced state-of-the-art results on several tasks \citep{dwivedi2020benchmarkgnns}, and demonstrate the inductive power to generalize the message passing mechanism to unseen nodes or even entirely new (sub)graphs \citep{hamilton2017inductive,velivckovic2018graph}. In \citep{appleby2020kriging}, spatial GNNs were applied to kriging task on a static graph. However, this work did not fully consider the temporal dependencies.

\subsection{Deep Learning for Missing Data Imputation}

Kriging applications are highly related to the missing data imputation problems. The former is about how to estimate values for unsampled
locations, while the latter is a process of filling in missing values for sampled locations. More recently, there has been a surge of interest in applying deep learning techniques to the missing data imputation problems. \citet{smieja2018processing} process the missing values of neural network inputs by simply replacing the neuron’s response in the first hidden layer with its expected value, and they also provide mathematical proof for the effectiveness of this strategy. This finding can be served as a theoretical foundation for masking the missing values in the inputs with a certain value. Several studies have developed various models including recurrent neural networks \citep{cui2020stacked}, generative adversarial networks \citep{yoon2018gain}, and variational autoencoders \citep{nazabal2020handling} to fill in the ``masked'' inputs. Those approaches ignore the local dependencies (e.g., similarities between adjacent locations). To fill this gap, some approaches have used GNNs to capture the local dependencies \citep{cui2020graph, spinelli2020missing}, they have shown the GNNs can effectively capture local dependencies in the presence of missing values. However, the GNNs based missing data imputation methods are limited in randomly missing cases within a set of sampled sensors. They can not be generalized to kriging tasks, where a new sensor set unseen during the training phase is present during the inference phase.

\section{Methodology}
\label{M:3}
\subsection{Problem Description}

\begin{figure*}[!ht]
\centering
\includegraphics[width = 0.6\textwidth]{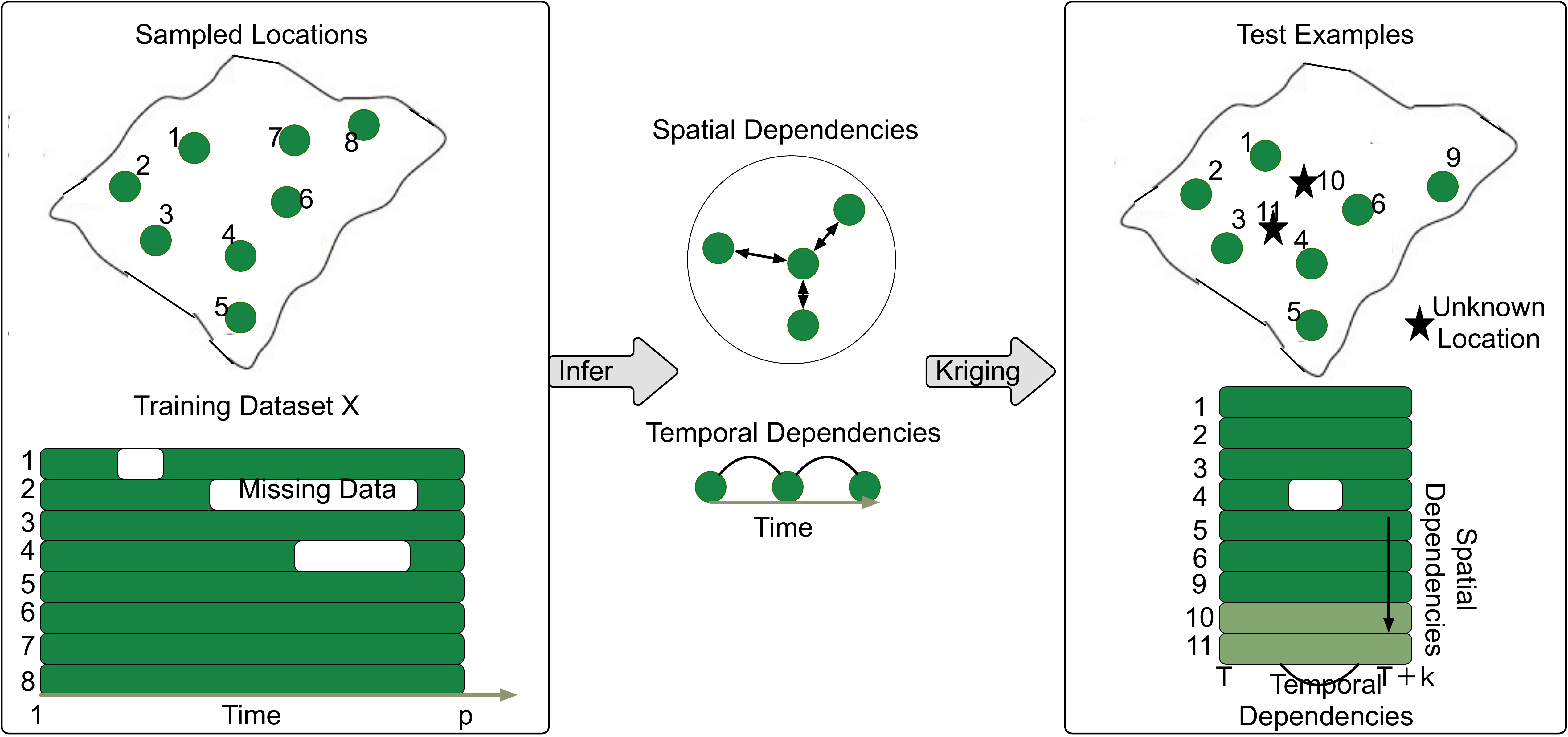}
\caption{Illustration of real-time kriging task, where the goal is to estimate the values of unsampled locations using the learned spatiotemporal dependencies. Note that the set of observedsensors during $[T, T+k]$ is not necessarily the same as the set during training time $[1, p]$. For example, during $[T, T+k]$, the sensors \{7,8\} are removed, and a new sensor \{9\} reported information.}
\label{Fig:2}
\end{figure*}

Our work focuses on the same real-time spatiotemporal kriging task as in \citep{wu2020inductive} (see Figure \ref{Fig:2}). Let $[t_1, t_2] = \{t_1, t_1 + 1, \ldots, t_{2} - 1, t_2\}$ denote a set of time points. Suppose we have data from $n$ sensors during a historical period $[1, p]$ ($n = 8$ in Figure \ref{Fig:2}, corresponding to sensors $\{1,\ldots,8\}$). Note that we use three terms---sensor, location and node---interchangeably throughout this paper. We denote the available training data by a multivariate time series matrix $X \in \mathbb{R}^{n \times p}$. Our goal is to infer a kriging model $\mathcal{G}(\cdot; \Lambda)$ based on $X$, with $\Lambda$ being the set of model parameters, which encode the spatiotemporal dependencies within the training data. Figure~\ref{Fig:2} also shows a test data example with $k$ time points. It should be noted that in our setting the kriging tasks could vary over time (or for each test sample) given sensor availability: some sensors might be not functional, some sensors may retire, and new sensors can also be introduced. Moreover, the number of interested time points $h$ can also vary over time. Taking the test sample in Figure~\ref{Fig:2} as an example; our spatiotemporal kriging task is to estimate the the signals $X^m_t \in \mathbb{R}^{n^m_t \times h_t}$ on $n^m_t$ unknown locations/sensors in black star (i.e., sensors \{10,11\}) based on observed data $X^o_t \in \mathbb{R}^{n^o_t  \times h_t}$ (in green) from $n^o_t$ sensors (i.e., a new sensor 9 emerged during kriging). The missing data also exist in the sampled locations. Obviously, the learned spatiotemporal dependencies in model $\mathcal{G}(\cdot; \Lambda)$ should be generalizable to unseen spatiotemporal points, which makes our problem more challenging than missing data imputation. Given the variation in data availability, we also prefer to have a model that is invariant to the size of the matrix $X^o_t$ and $X^m_t$.

\begin{figure}[!ht]
\centering
\includegraphics[width = 0.45\textwidth]{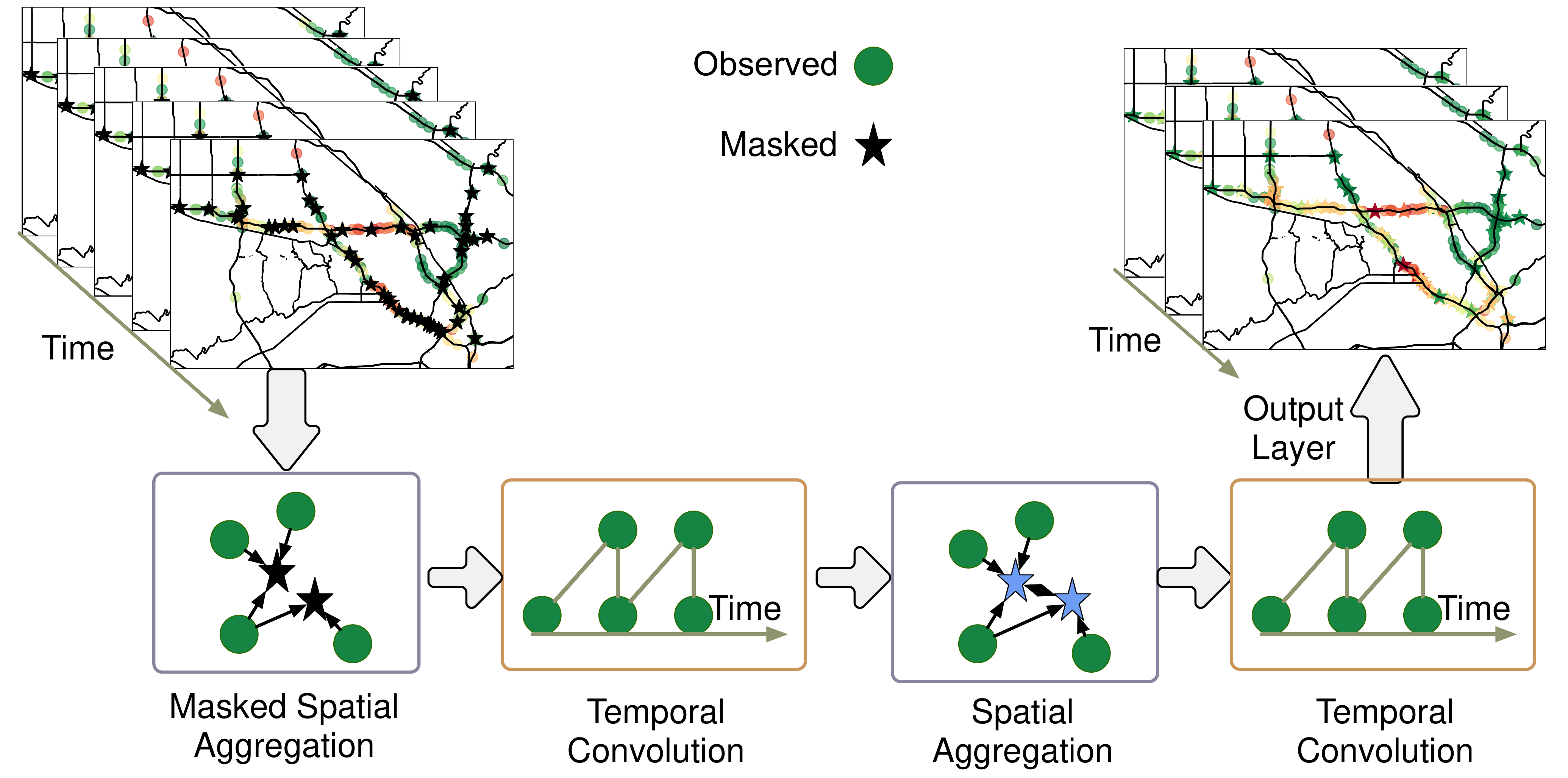}
\caption{An example of SATCN on traffic speed datasets. The color of the marks represents the reading value of the sensors. The given SATCN consists of two spatial aggregation blocks and two temporal convolution blocks.}
\label{Fig:1}
\end{figure}

\subsection{Spatial Aggregation and Temporal Convolution Networks (SATCN)}

We introduce SATCN---a novel deep learning architecture---as the spatiotemporal kriging model $\mathcal{G}(\cdot; \Lambda)$, where $\Lambda$ is the set of parameters. We design SATCN to generate the estimation of $\hat{X}_t^m$ based on input $X_t^o$ and additional information of the underlying graph, i.e., $\hat{X}_t^m = \mathcal{G}\left(X_t^o,\cdots;\Lambda\right)$. The proposed SATCN consists of two basic building blocks: spatial aggregation networks (SANs) and temporal convolutional networks (TCNs) in alternating order (see Figure~\ref{Fig:1}). The input size and output size of SATCN are ${n_t  \times (h_t + u)}$ and ${n_t  \times h_t}$, respectively, where $u$ means the size reduction as a result of TCN (e.g., if we use two TCNs with a width-$w_t$ kernel in SATCN, the size reduction will be $u = 2(w_t - 1)$). The spatial aggregation layer is built on a special graph neural network---principal neighborhood aggregation (PNA) \citep{corso2020principal}. Note that the input of SATCN also contains the unknown sensors on which we will perform kriging (see the \textit{masked spatial aggregation} layer in Figure~\ref{Fig:1}); however, we propose a masking strategy to forbid the unknown locations to send messages to their neighbors. We next introduce the details of SATCN. We summarize some important notations related with SATCN definition and the masking strategy in Table~\ref{tab:notation}.

\begin{table}[!t]
\caption{Key notations.}
\label{tab:notation}
\centering
\begin{tabular}{lp{7cm}}
\toprule
Symbol & Description \\
\midrule
$n_m$ & number of simulated unknown locations \\
$k$ & number of nearest neighbor for message passing \\
$h$ & the time length of kriging target \\
$u$ & the time length reduction caused by temporal convolution \\
$A_s$ & the masked adjacency matrix of the $s$-th sample for the first spatial aggregation layer \\
$\hat{A}_s$ & the adjacency matrix of the $s$-th sample for the higher spatial aggregation layer \\
$S$ & batch size for sampling and training \\
$w_l$ & the kernel length of the $l$-th temporal convolution layer\\
\bottomrule
\end{tabular}
\end{table}

\subsection{Training Sample and Adjacency Matrix Construction}

\begin{figure}[!t]
\centering
\includegraphics[width = 0.48\textwidth]{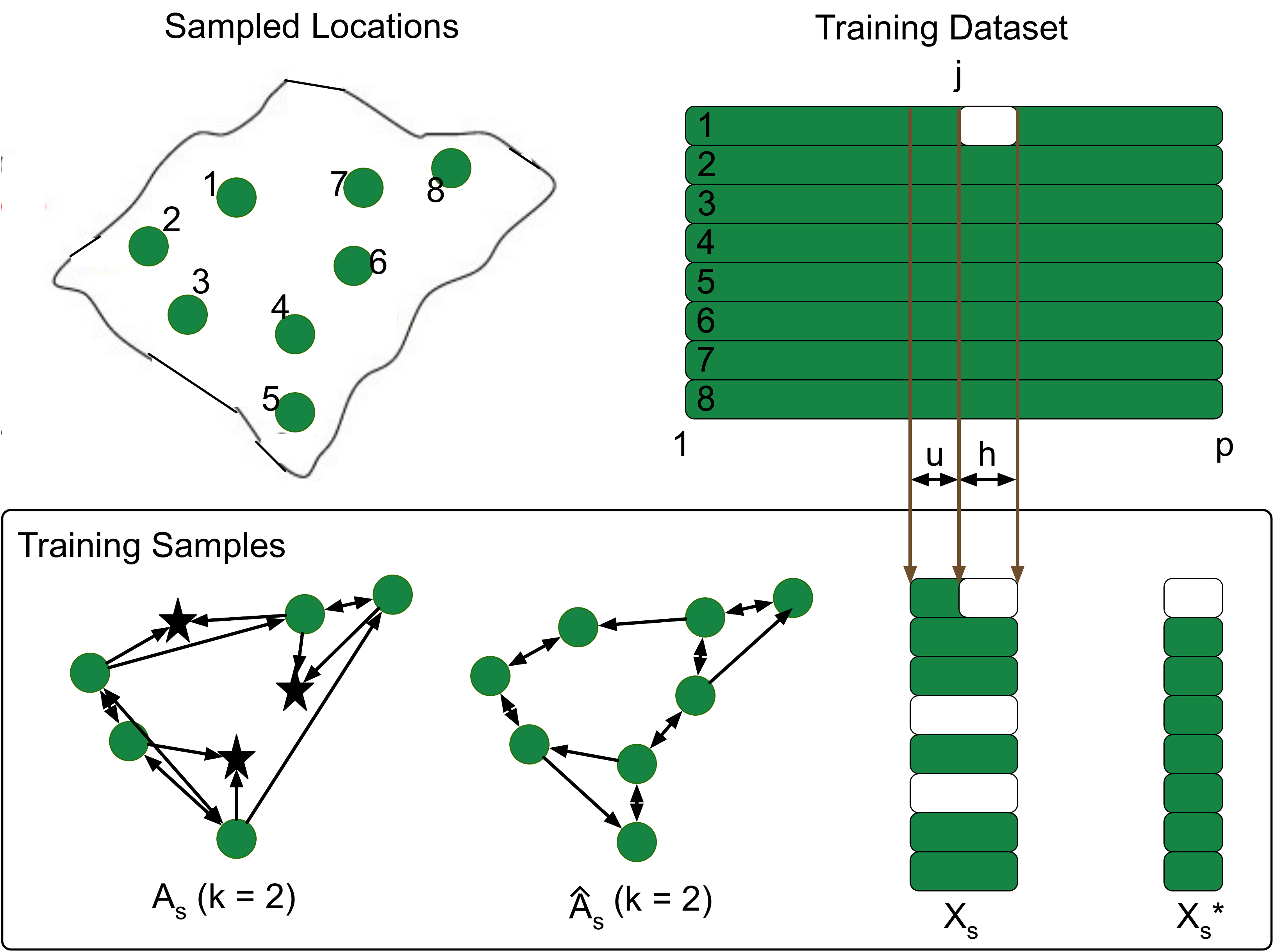}
\caption{An example of training sample $X_s$, $X^*_s$, $A_s$, and $\hat{A}_s$ from Algorithm. \ref{alg:A1}. In this example, the nearest neighbor $k$ is set to 2 to generate $A$ and $\hat{A}$, The number $n_m$ is set to 2. The sensors $\{4,6\}$ are randomly picked up as masked sensors, they can not send message to other locations in $A_s$, and their values are set to 0 in $X_s$.}
\label{Fig:sample}
\end{figure}

Our first step is to prepare training samples from the historical dataset $X$. The random sampling procedure is given in Algorithm~\ref{alg:A1}. The key idea is to randomly simulate $n_m$ unknown locations among the $n$ observed locations. We also show an example of training sample generation in Figure~\ref{Fig:sample}.
As SATCN uses GNNs to capture the spatial dependencies, we construct a graph over locations/sensors for training. In such a graph, the $n_m$ unknown locations cannot pass messages to their neighbors. We define the adjacency matrix $A$ of this graph according to the following rule:
\begin{equation}
    A_{ij} = \begin{cases}
    1 - \frac{dist(s_i, s_j)}{d_{max}},     & \text{if } i \in \bar{\Omega}_{\mathcal{K}_j},\\
    0,  &  \text{otherwise},
    \end{cases}
    \label{adj}
\end{equation}
where $\Omega$ is the set of unknown sensors with $|\Omega| = n_m$, $\bar{\Omega}_{\mathcal{K}_j}$ is the set of $k$-nearest neighbors for the $j$-th sensor in known set $\bar{\Omega}$, $dist(s_i, s_j)$ is the distance between the sensors $i$ and $j$, and $d_{max}$ is the maximum distance between any two sensors in the training data.  In some applications, the missing locations are evolving with time (e.g., some observations from satellites are obscured by clouds). To deal with those cases, the adjacency matrix $A$ should be time-evolving, that is, the locations with missing data are always forbidden to send messages to their neighbors. We also set the values of masking locations to $0$, ensuring the model has no access to unknown observations; however, the set values have no impact on our model as the masking locations have been forbidden to send messages. Considering that SATCN contains multiple spatial aggregation layers, we expect the unknown sensors to also generate meaningful information after the aggregation in the first layer. Therefore, we define the adjacency matrix $\hat{A}$ of the subsequent SAN layer as:
\begin{equation}
    \hat{A}_{ij} = \begin{cases}
    1 - \frac{dist(s_i, s_j)}{d_{max}},     & \text{if } i \in {\mathcal{K}_j},\\
    0,  &  \text{otherwise},
    \end{cases}
    \label{adj_}
\end{equation}
where ${\mathcal{K}_j}$ is the set of $k$-nearest neighbors for the $j$-th sensor.

\begin{algorithm}[!t]
\caption{Training sample generation}
\label{alg:A1}
\begin{algorithmic}[1]
\REQUIRE Historical data $X$ from $n$ observed sensors/locations over period $[1,p]$ (size $n\times p$). \\ Parameters: window length $h$, SATCN temporal reduction $u$, sample size $S$ in each iteration, and the maximum number of iterations $I_{\max}$, number of simulated unknown locations $n_m$. \\
\FOR {$\text{iteration} = 1:I_{\max}$}
\FOR {$\text{sample} = 1:S$}
\STATE Randomly choose a time point $j$ within range $\left[1+u, p-h\right]$. Let $J_{\text{sample}} = \left[j-u,j+h\right)$, $J^*_{\text{sample}} = \left[j,j+h\right)$.
\STATE Obtain submatrix signal $X_{\text{sample}} = X[:, J_{\text{sample}}]$ with size of $n \times (h + u)$, $X^*_{\text{sample}} = X[:, J^*_{\text{sample}}]$ with size of $n \times h$.
\STATE Generate a random set $\Omega_s$ with size $n_m$ (number of nodes selected as missing) with $n_m \leq n$.
\STATE Let $X_{\text{sample}}[\Omega_s, :] = 0$ to ensure the model has no access to unknown observations.
\STATE Construct adjacency matrix $A_{\text{sample}} \in \mathbb{R}^{n \times n}$ and $\hat{A}_{\text{sample}} \in \mathbb{R}^{n \times n}$ based on Eqs.~\eqref{adj} and \eqref{adj_}.
\ENDFOR
\STATE Use sets $\{X_{1:S}\}$, $\{X^*_{1:S}\}$, $\{A_{{1:S}}\}$, $\{\hat{A}_{{1:S}}\}$ to train SATCN.
\ENDFOR
\end{algorithmic}
\end{algorithm}

To train the model, we use all spatiotemporal points in $X^*_{1:S}$ as test data. We use the mean absolute error (MAE) term to measure the reconstruction loss between output $\mathcal{G}(X, A, \hat{A}; \Lambda)$ and all observed true labels $X^*_{1:S}$. The learning objective of SATCN is
\begin{equation}
    \min_\Lambda \frac{1}{S} \sum^S_{s = 1} \|\mathcal{G}(X_s, A_s, \hat{A}_s; \Lambda) -  X^*_s \|_1.
    \label{loss}
\end{equation}

\subsection{Spatial Aggregation Network (SAN)}

Based on the constructed adjacency matrices $A_s$ and $\hat{A}_s$, the goal of performing spatial aggregation is to capture the distribution of messages that a sensor receives from its neighbors. Most existing approaches use a single aggregation method, with $mean$ and $weighted$ $sum$ being the most popular choices in spatiotemporal applications (see e.g., \citep{KipfW17,xu2018powerful}). However, the single aggregation method has two limitations: 1) in many cases, the single aggregation fails to recognize the small difference between different messages \citep{corso2020principal,dehmamy2019understanding}; and 2) the true distances to neighbor locations have a large impact on kriging, but existing GNNs do not fully consider the effect of distances. Therefore, for the spatiotemporal kriging task, it is necessary to introduce distance-sensitive aggregators in SATCN. Inspired by the recent work of principal neighborhood aggregation proposed by \citet{corso2020principal}, we leverage multiple aggregators on nodes and edges (links) to capture the spatial dependencies. Specifically, we introduce the following aggregators to SATCN:
\begin{itemize}
    \item {\textbf{Mean:}} $\mu_i(x^l) = \frac{1}{k} \sum_{A^l_{ji} > 0} x^l_j$,
    \item {\textbf{Weighted Mean:}} $\hat{\mu}_i(x^l) = \frac{\sum_{A^l_{ji} > 0} A^l_{ji} x^l_j}{\sum_{A^l_{ji} > 0} A^l_{ji}} $,
    \item {\textbf{Softmax:}} $\text{softmax}_i(x^l) = \sum_{A^l_{ji} > 0} \frac{x^l_j exp(x^l_j)}{\sum_{A^l_{ki} > 0}{exp(x^l_k)}}$,
    \item {\textbf{Softmin:}} $\text{softmin}_i(x^l) = -\text{softmax}_i(-x_i^l)$,
    \item {\textbf{Standard deviation:}}

    $\sigma_i(x^l)=\sqrt{RELU\left(\mu_i\left(\left({x^l}\right)^2\right) - \mu_i\left({x^l}\right)^2\right) + \epsilon}$,
    \item {\textbf{Mean distance:}} $\mu^d_i(A^l) = \frac{1}{k} \sum_{A^l_{ji} > 0} A^l_{ji}$;
    \item {\textbf{Standard distance deviation:}}

    $\sigma^d_i(A^l)=\sqrt{RELU\left(\mu^d_i\left(\left({A^l_{:i}}\right)^2\right) - \mu^d_i\left({A^l_{:i}}\right)^2\right) + \epsilon}$,
\end{itemize}
where $x^l$ denotes the feature vector of node $i$ in the $l$-th layer, $A^l_{ji}$ denotes the distance from $j$ to $i$ given by the $l$-th layer adjacency matrix, $A^l_{ji}>0$ means that there exists message passing flow from $j$ to $i$, and $\epsilon$ in standard deviation is a small positive value to ensure $\sigma$ is differentiable.

{{Mean aggregation}} treats all neighbors of a node equally without considering the effect of distance. {{Weighted mean aggregation}} takes distance weights $A^l_{ji}$ into consideration. {{Softmax  aggregation}} and {{Softmin aggregation}} give indirect measures  for the maximum and minimum value of the received messages, which offer more generalization capability of GNN \citep{velivckovic2019neural}. The aggregations above are essentially the same as in \citep{corso2020principal}, capturing the variation of features that one node receives from its neighbors. In addition, we include {{Mean distance aggregation}} and {{Standard deviation distance aggregation}}, which characterize the distribution of spatial distances from a certain node to the neighboring nodes. To make GNNs better capture the effect of spatial distance, we suggest further adding logarithmic (amplification) and inverse logarithmic (attenuation) scalers \citep{xu2018powerful, corso2020principal}: \begin{equation}
\begin{aligned}
S_i^{amp}\left(agg_i \right) = & \frac{\left(\log{\left(\sum{A^l_{:i}}  + 1\right)}\right) \cdot agg_i}{deg}, \\
S_i^{att}\left(agg_i \right) = & \frac{deg \cdot agg_i}{\left(\log{\left(\sum{A^l_{:i}}  + 1\right)}\right) },
\end{aligned}
\end{equation}
where $deg$ equals to $ \frac{1}{|train|}\sum_{i \in train} \left(\log{\left(\sum{\hat{A}_{:i}} + 1\right)}\right)$, $\hat{A}$ is the adjacency matrix constructed by all training locations using the rule in Eq.~\eqref{adj_}.

\begin{figure}[!t]
\centering
\includegraphics[width = 0.45\textwidth]{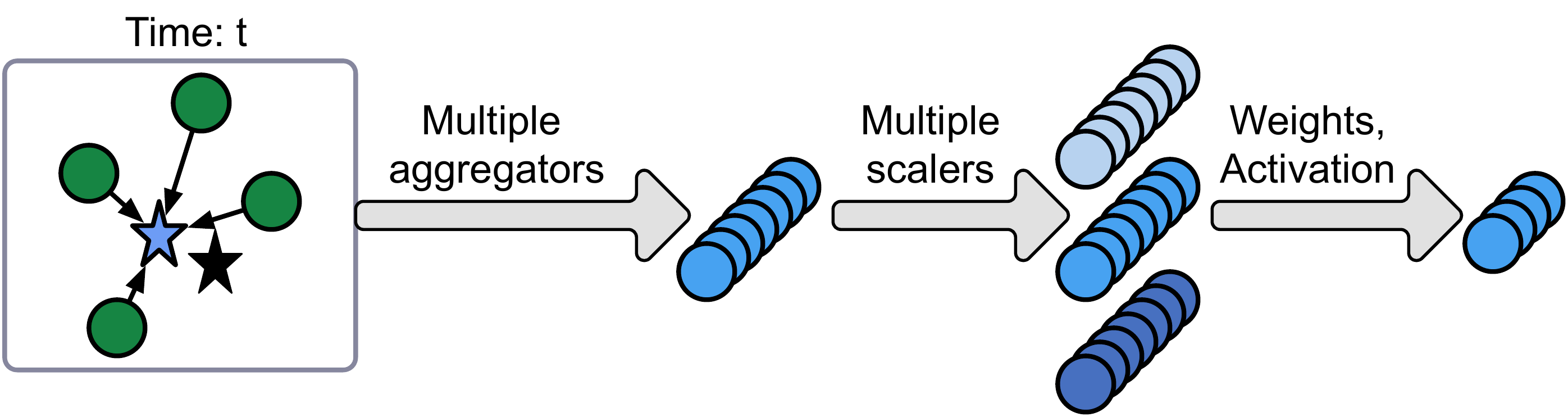}
\caption{The masked spatial aggregation network. In this figure, an unknown star node colored in black is the nearest neighbor of the blue star node, but we prevent it from passing messages to the blue star. After the message aggregation and scale in SAN, the blue star receive meaningful information. To visualize this process, we transform its star shape to the circle shape. }
\label{Fig:3}
\end{figure}

We then combine the aggregators and scalers using the tensor product $\otimes$:
\begin{equation}
\label{agg_sca}
    \bigoplus = \begin{bmatrix}
       I \\
       S^{amp} \\
       S^{att}
     \end{bmatrix}  \otimes \begin{bmatrix}
       \mu \\
       \hat{\mu} \\
       \text{softmax}   \\
       \text{softmin}  \\
       \sigma \\
       \mu^d \\
       \sigma^d
     \end{bmatrix},
\end{equation}
where $I$ is an identity matrix, $\otimes$ is to multiply all scalers and aggregators together and then stack them on top of each other. We add weights and activation function to $\bigoplus x$ obtaining SAN:
\begin{equation}
    X_{:t}^{l+1} = f^l\left(\Phi^l \bigoplus X_{:t}^l + b_s^l\right),
\end{equation}
where $X_{:t}^l \in \mathbb{R}^{n \times c_l}$ is the $l$-th layer output at $t$-th time point, $\Phi^l \in \mathbb{R}^{c_{l+1} \times n_{sc}n_{ag}c_l}$ is the $l$-th layer weights, $b_s^l \in \mathbb{R}^{c_{l+1}}$ is the bias, $n_{sc}$ is the number of scalers, $n_{ag}$ is the number of aggregators  and $f^l$ is the activation function. For each time point $t$, the equal spatial aggregation operation with the same weights $\Phi^l$ and $b^l$ is imposed on $X^l_{:t} \in \mathbb{R}^{n \times c_l}$ in parallel. By this means, the SAN can
be generalized in 3D spatiotemporal variables, denoted as ``$\Phi^l \oplus_{S} X^l$''. For example, the inputs $X^1$ of SATCN is with size of $n \times (h + u)$, $\bigoplus X_{:t}^1$ will result in a $n \times (h + u) \times 21$ tensor. Given a $\Phi^1 \in \mathbb{R}^{c_2 \times 21}$. $\Phi^1 \oplus_{S} X^1$ will result in a tensor with size $n \times (h + u) \times c_2$. We illustrate the operation for the Masked SAN (the first layer of SATCN) in Figure \ref{Fig:3}.

\subsection{Temporal Convolutional Network (TCN)}

To capture temporal dependencies, we take advantage of the
temporal convolution networks (TCNs). The advantage of TCNs on the kriging task includes: 1) TCNs can deal with input sequences with varying length, allowing us to perform kriging for test samples with different numbers of time points; 2) TCNs are superior to recurrent neural networks (RNN) with fast training and lightweight architecture \citep{gehring2017convolutional}.  A shared gated 1D convolution is applied on each node along the temporal dimension. A width-$w_l$ TCN passes messages from $w_l$ neighbors to the target time point. Note that we do not use padding in our TCN, and thus a width-$w_l$ TCN will shorten the temporal length with number of $w_l - 1$. TCN maps the input $X^l \in \mathbb{R}^{n \times h_l \times c_l}$ to a tensor
\begin{equation}
    T^l = \Gamma^l \ast X^l + b_t^l,
\end{equation}
where $T^l \in \mathbb{R}^{n \times (h_l-w_l+1) \times c_{l+1}}$. $\ast$ is the shared 1D convolutional operation, $\Gamma^l \in \mathbb{R}^{w_l \times c_l \times c_{l+1}}$ is the convolution kernel. A benefit brought by TCN is that our model is not dependent on the length of training/test time period. In our model, data of each time point is only related to other points within a receptive frame determined by $w_l$. The length of temporal window has no impact on our model as time points beyond the receptive frame cannot pass information to the target time point.

\section{Experiments}
\label{E:4}

In this section, we conduct experiments on several real-world spatiotemporal datasets to evaluate the performance of SATCN.

\subsection{Experiment Setup}
\subsubsection{Dataset Descriptions.}
We evaluate the performance of SATCN on six real-world datasets in diverse settings: (1) \textbf{METR-LA}\footnote{\url{https://github.com/liyaguang/DCRNN}} is a 5-min traffic speed dataset collected from 207 highway sensors in Los Angeles, from Mar 1, 2012 to Jun 30, 2012. (2) \textbf{MODIS}\footnote{\url{https://modis.gsfc.nasa.gov/data/}} consists of daily land surface temperatures measured by the Terra platform on the MODIS satellite with 3255 down-sampled grids from Jan 1, 2019 to Jan 16, 2021. It is automatically collected by \textit{MODIStsp} package in \textit{R}. This dataset poses a challenging task as it contains 39.6\% missing data. (3) \textbf{USHCN}\footnote{\url{https://www.ncdc.noaa.gov/ushcn/introduction}} consists of monthly precipitation of 1218 stations in the US from 1899 to 2019. The temporal resolutions of METR-LA is 5-min. With regards to spatial distance, we compute the pairwise haversine distance matrices for MODIS and USHCN; METR-LA uses travel distance on the road to determine the reachability among sensors in the transportation network. For METR-LA and USHCN, we test 4 cases: (a) \textbf{7T8S}: The first 70\% (in time) and approximately 80\% locations/sensors (in space) are used for training, the other 20\% sensors during the last 30\% time periods are used for evaluation. (b) \textbf{5T5S}: The first 50\% and approximately 50\% locations/sensors are used for training, the other 50\% sensors during the last 50\% time periods are used for evaluation. (c) \textbf{7T8S3M}: We additionally add 30\% missing data to the observations (80\% sampled sensors) under case 7T8S. (d) \textbf{5T5S5M}: We additionally add 50\% missing data to the observations under case 5T5S. For all cases, 10\% sampled sensors are used as validation data. The MODIS dataset itself contains missing data, we only evaluate our model on case 7T8S and 5T5S for MODIS.

\subsubsection{Baseline Models.}

We choose both traditional kriging models and state-of-the-art deep learning models as our baseline models. The group of traditional models includes: (1) \textbf{kNN}: K-nearest neighbors, which estimates the signal on unknown sensors by taking the average values of \textit{K} nearest sensors in the network. (2) \textbf{OKriging}: ordinary kriging \citep{cressie1988spatial}, which corresponds to a Gaussian process regression model with a global mean parameter. We implement OKriging with the \textit{autoKrige} function in \textit{R} package \textit{automap}. The OKriging method only uses spatial dependencies. We tried to implement a spatiotemporal kriging baseline via the \textit{R} package \textit{gstat}. However, learning a proper spatiotemporal variogram is very challenging give the large size of the datasets. Thus, we did not include a spatiotemporal kriging baseline in this work. (3) \textbf{GLTL}: Greedy Low-rank Tensor Learning, a transductive tensor factorization model for spatiotemporal co-kriging \citep{bahadori2014fast}. GLTL can handle the co-kriging problem with multiple variables (e.g., $location\times time \times variable$). We reduce GLTL into a matrix version, as our task only involves one variable for all the datasets. In addition, we compare SATCN with the following state-of-the-art deep-learning approaches: (4) \textbf{IGNNK}: Inductive Graph Neural Network for kriging \citep{wu2020inductive}, an inductive model that combines dynamic subgraph sampling techniques and a spectral GNN \citep{li2018diffusion} for the spatiotemporal kriging task.
 We use a Gaussian kernel to construct the adjacency matrix for GNN as in \citep{wu2020inductive}:
\begin{equation}
    A_{ij} = \exp\left(-\left(\frac{\text{dist}\left(v_i , v_j\right)}{\sigma}\right)^2\right),
\end{equation}
 where $A_{ij}$ stands for adjacency or closeness between sensors $v_i$ and $v_j$, $\text{dist}\left(v_i , v_j\right)$ is the distance between $v_i$ and $v_j$, and $\sigma$ is a normalization parameter. This adjacency matrix can be considered a squared-exponential (SE) kernel (Gaussian process) with $\sigma/\sqrt{2}$ as the lengthscale parameter. Different from \citep{wu2020inductive} which chooses $\sigma$ empirically, we first build a Gaussian process regression model based on the training data from one time point and estimate the lengthscale hyperparemter $l$, and then define $\sigma=\sqrt{2}l$. We find that this procedure improves the performance of IGNNK substantially compared with \citep{wu2020inductive}. (5) \textbf{KCN-Sage}: In \citep{appleby2020kriging}, several GNN structures are proposed for kriging tasks, and we use KCN-Sage based on GraphSAGE \citep{hamilton2017inductive} as it achieves the best performance in \citep{appleby2020kriging}. The original KCN models cannot be directly applied under the inductive settings. To adapt KCNs to our experiments, we use Eq.~\eqref{adj_} to construct the adjacency matrices of KCN-Sage and Algorithm~\ref{alg:A1} to train the model.

\subsubsection{Evaluation Metrics}

We measure model performance using the following three metrics:
\begin{equation}
    \text{RMSE} = \sqrt{ \frac{1}{|N|} \sum_{i\in N} (x_i - \hat{x})^2},
\end{equation}
\begin{equation}
    \text{MAE}=\frac{1}{|N|} \sum_{i\in N} \left|x_i - \hat{x}\right|,
    \end{equation}
where $x_i$ and $\hat{x}_{i}$ are the true value and estimation, respectively, and $\bar{x}$ is the mean value of the data.

\begin{table*}[!t]
\caption{Kriging performance (in RMSE/MAE) comparison of different models on three datasets.}
\label{tab:comparison}
\centering
\begin{tabular}{c|c|c c c c c c}
\toprule
Data & Case & SATCN & KCN-SAGE & IGNNK & GLTL & Okriging & kNN \\
\midrule
\multirow{4}{*}{METR-LA} & 7T8S &\textbf{8.35/5.40}  &8.52/5.61  & 9.08/5.82 & 11.04/7.55 & -/- & 11.62/6.33 \\
& 5T5S &\textbf{9.83/6.38}  &10.29/6.54  & 11.22/7.12 & -/- & -/- & 12.79/7.04\\
& 7T8S3M &\textbf{9.07/5.72}  &9.22/5.99  & 9.93/6.45 & 12.48/8.47 &  -/- & 11.84/6.79\\
& 5T5S5M &\textbf{10.23/6.65} &10.61/6.78  & 11.59/7.70 & -/- & -/- & 12.88/7.46\\
\midrule
\multirow{4}{*}{USHCN} & 7T8S &\textbf{3.54/2.14} &3.56/2.18  & 3.62/2.22 & 4.12/2.67 & 3.73/2.26 & 3.76/2.27\\
& 5T5S &\textbf{3.71/2.25}  &3.78/2.35  & 3.85/2.37 & 4.31/2.81 & 3.86/2.37  & 3.87/2.41\\
& 7T8S3M &\textbf{3.67/2.22}  &4.12/2.58  & 4.49/2.73 & 4.14/2.69 &  3.99/2.47 & 4.50/2.74\\
& 5T5S5M &\textbf{4.03/2.45} &4.84/3.12  & 6.02/3.89 & 4.76/3.12 & 4.42/2.80 & 5.52/3.60\\
\midrule
\multirow{2}{*}{MODIS} & 7T8S &\textbf{1.38/0.94} &1.42/0.95  &-/- &-/- & 1.62/1.09 &1.70/1.14  \\
& 5T5S &\textbf{1.54/1.01}  &1.58/1.05 & -/- & -/- &  4.89/3.02 & 4.90/3.18\\
\bottomrule
\end{tabular}
\end{table*}


\subsection{Overall Performance}
\subsubsection{Performance Comparison.} In Table. \ref{tab:comparison}, we present the results of SATCN and all baselines on six datasets. As the spatial relationship of METR-LA is determined by road reachability, we cannot apply OKriging--which directly defines locations in a geospatial coordinate---on these two datasets. As can be seen, the proposed SATCN consistently outperforms other baseline models, providing the lowest error for almost all datasets under all cases. SATCN generally outperforms the spectral-based GNN counterparts--- IGNNK on those datasets, we also observe that SATCN and KCN-Sage take less samples to converge compared with the spectral approach IGNNK. It indicates that the spatial GNNs are more suitable for inductive tasks compared with the spectral ones.

Another interesting finding from Table~\ref{tab:comparison} is that SATCN performs well on MODIS dataset and 5T5S5M cases, which contain high ratio missing data in the observed samples. IGNNK and GLTL sometimes fail to work under those conditions because that data shows substantial spatially correlated corruptions (i.e., the cloud). SATCN is robust to missing data as it ensures that every location has $\mathcal{K}$ observable neighbors using the adjacency matrix construction rule given in Eq.~\eqref{adj}.

As expected, SATCN is more advantageous compared to other baselines when the missing ratio is higher. SATCN significantly outperforms other baselines under \textbf{5T5S5M} for USHCN dataset. Its advantages may be attributed to the utilization of TCN, since temporal dependencies are more important when fewer nodes' information is observable.

\begin{figure}[!t]
    \centering
    \includegraphics{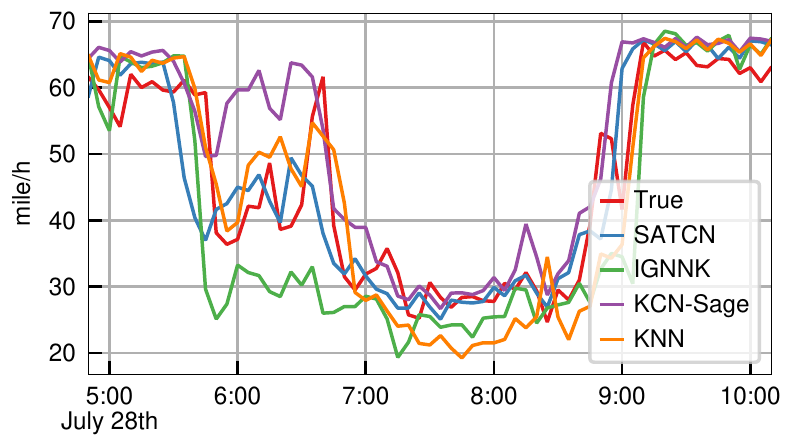}
    \caption{Kriging performance on an unknown node in METR-LA dataset in July 28th, 2012.}
    \label{fig:satcn_metr}
\end{figure}

\begin{figure}[!t]
    \centering
    \includegraphics{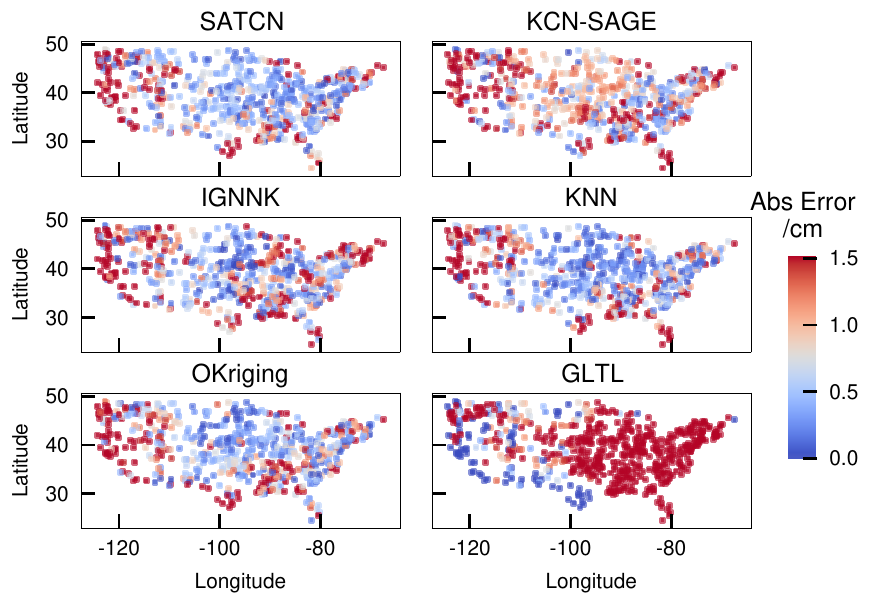}
    \caption{Absolute kriging errors on test locations in January, 1981 of USHCN dataset. Only a third of the sensors are observed.}
    \label{fig:udata_result}
\end{figure}

\begin{figure}[!ht]
    \centering
    \includegraphics{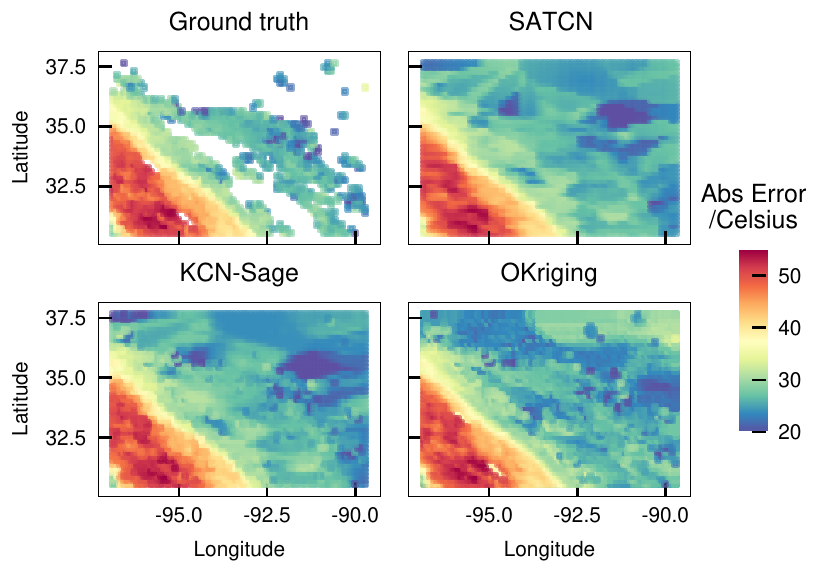}
    \caption{Kriging results in Jun 8th, 2020 of the MODIS dataset. Land temperature ground truth and the results of SATCN, KCN-Sage, and OKriging, respectively.}
    \label{fig:modis_cloud}
\end{figure}

Figure~\ref{fig:satcn_metr} gives the temporal visualization of kriging results for one sensor from METR-LA dataset under \textbf{5T5S} case. It is clear that SATCN model produces the closest estimation toward true values. With the learned temporal dependencies, SATCN can better approximate the sudden speed drop of morning peak during 5:00 AM-10:00 AM due to the benefits from the temporal dependencies. We also visualize the results on USHCN dataset under \textbf{5T5S5M} case in Figure \ref{fig:udata_result}, it is obvious that SATCN outperforms other methods in this sparsely observed case. To qualitatively illustrate the performance of SATCN under missing data, we also visualize the interpolation results on the areas covered by clouds in Figure~\ref{fig:modis_cloud}. From the results we can find that SATCN gives a more physically consistent predictions for areas covered by clouds. Results of KCN-SAGE and KNN contain more small-scaled anomalies because they do not take advantage of temporal and distance information. OKriging generates over-smoothed predictions, which are not consistent with the observations of known areas.

\subsubsection{Parameter Sensitivity Analysis:} SATCN has many parameter settings, the key parameters include the number of neighbors for message passing, the TCN kernel length (temporal receptive field), the choice of aggregators and scalers, number of spatial aggregation and temporal convolution layers, the number of simulated unknown nodes, and the number of hidden neurons.

\begin{figure*}[!ht]
\centering
\subfigure[Errorbars with respect to the number of nearest neighbors]{
\includegraphics{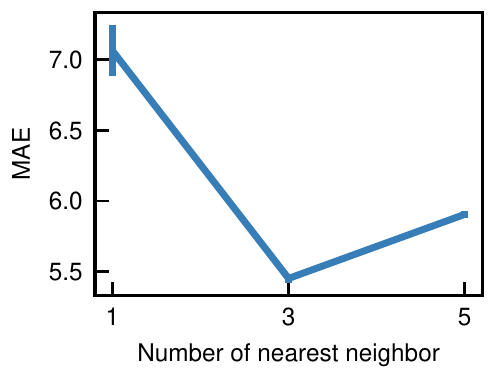}
\label{fig_ek}}
\subfigure[Errorbars with respect to the temporal receptive field]{
\includegraphics{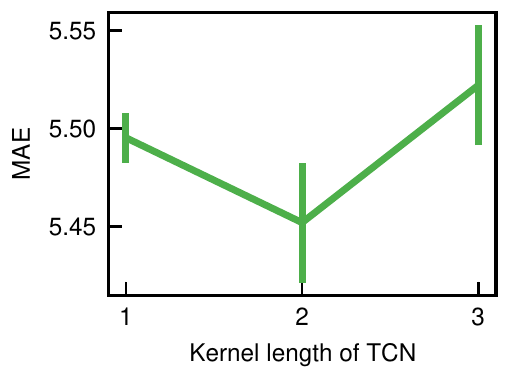}
\label{fig_ew}}
\subfigure[Errorbars with respect to the setting of aggregators and scalers]{ \includegraphics{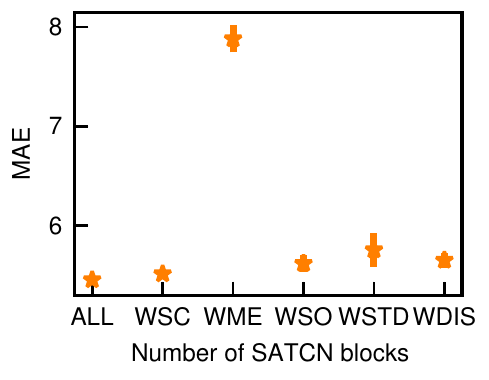}
\label{fig_ea}}

\subfigure[Errorbars with respect to the number of SATCN layers]{
\includegraphics{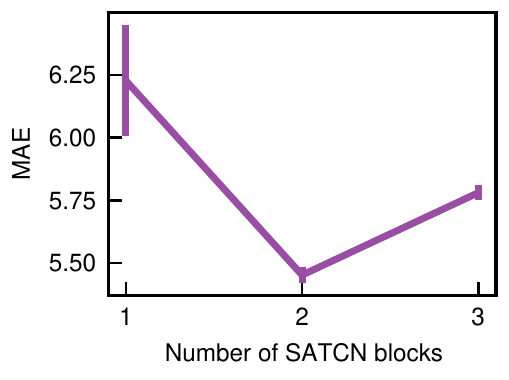}
\label{fig_el}}
\subfigure[Errorbars with respect to the number of hidden neurons]{
\includegraphics{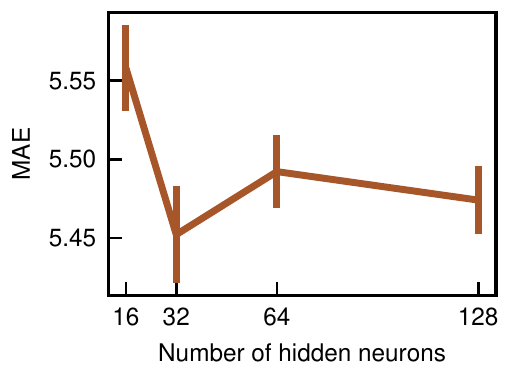}
\label{fig_en}}
\subfigure[Errorbars with respect to the number of simulated unknown nodes]{ \includegraphics{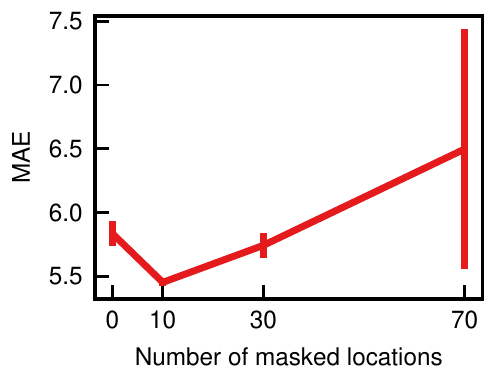}
\label{fig_em}}
    \caption{SATCN performance under different settings.}
\end{figure*}

To evaluate the impact of $k$, we fix TCN kernel length $w_t$ to 2, the number of hidden neurons $c_l$ to 32, the number of simulated unknown locations $n_m$ to 10 for METR-LA 7T8S case, and vary $k$ in $\{1, 3, 5\}$. The results (mean value and standard deviation of last 8 steps after convergence) are reported in Figure~\ref{fig_ek}. The models with three neighbors achieve the lowest MAE. We speculate that traffic data can only propagate in a small spatial range. The correlation between close locations is strong but diminishes quickly when the distance increases. To evaluate the impact of temporal kernel length $w$, we fix $k$ to 3, and vary $w$ from 1 to 3. The results are given in Figure~\ref{fig_ek}. Compared with $k$, $w$ only affects SATCN marginally. Varying the temporal receptive field only has a little impact on SATCN. Surprisingly, the model with $w = 1$ also performs well, and it gives a relatively lower deviation.  A potential reason is that the strong spatial consistency alone could be sufficient to support the spatiotemporal kriging task. In other words, we can achieve a reasonably good result by performing kriging on each time snapshot.

The spatial aggregation network contains numerous aggregators and scalers. To distinguish their contribution, we also study the effects of different aggregator-scaler combinations (see Figure \ref{fig_ea}). We evaluate several models: \textbf{ALL} denotes the model with all aggregators and scalers in Eq~\eqref{agg_sca}; \textbf{WSC} denotes the model without logarithmic and inverse logarithmic scalers; \textbf{WME} denotes the model without mean related aggregators $\hat{\mu}$ and $\mu$; \textbf{WSO} denotes the model without softmax and softmin aggregators; \textbf{WSTD} denotes the model without standard deviation aggregator $\sigma$; \textbf{WDIS} denotes the model without distance-related aggregators $\hat{\mu}$, $\mu^d$ and $\sigma^d$. In Figure \ref{fig_ea}, \textbf{WME} gives the worst performance. It suggests that the mean and normalized mean values of the neighbors contain the most important information for kriging. The models \textbf{WSC}, \textbf{WSTD}, \textbf{WSO} and \textbf{WDiS} all give worse performance than the model with all aggregators and scalers. It suggests that using multiple aggregators and scalers can improve the kriging performance. In particular, the model \textbf{WSTD} gives relatively worse performance compared with models except \textbf{WME}, and its error deviation is larger than other models. It indicates that the message passing information within standard deviation aggregators can make the kriging model easier to learn. The logarithmic and inverse logarithmic scalers have the smallest impact on model accuracy. However, they contain some global information in the training graph, as they are based on the average weighted degree of the graph constructed by all training locations.

The expressiveness of graph neural networks can be theoretically improved by increasing the depth and width of GNNs \citep{loukas2019graph}. However, there are several factors that impede deeper and wider GNNs to perform better \citep{rong2019dropedge}. To study the effect of SATCN depth, we fix TCN kernel length $w_t$ to 2, the number of nearest neighbor $k$ to 3, the number of simulated unknown locations $n_m$ to 10, and vary the number of SATCN blocks in $\{1, 2, 3\}$ with hidden neuron fixed to $32$ in Figure~\ref{fig_el}. First, the SATCN model with only one layer is very difficult to converge, and give higher MAEs compared with deeper competitors. In SATCN, the first layer only utilizes the masked adjacency matrix $A$, which may lose some important information. Therefore the model with only one SATCN block is not sufficient to perform accurate kriging. Second, the model with three SATCN blocks are worse than the one with two blocks. It might be caused by the fact that the target dataset itself is locally correlated. A deeper SATCN will enlarge the spatiotemporal receptive frame of every unknown location, and thus incorporate unrelated information to the kriging model. To study the width of the SATCN model, we fix the depth of SATCN to 2, and vary the number of hidden neurons $c$ in $\{16, 32, 64, 128\}$ in Figure~\ref{fig_el}. We observe a large improvement when $c$ reaches 32. However, there is no evidence that the models with 64 and 128 hidden neurons are better than the one with 32 neurons. We only observe that the models with larger $c$ exhibits lower variance on test MAEs.

A key element of SATCN is the training sample generation algorithm in Algorithm~\ref{alg:A1}, which generates random graph structure by randomly masking. We also study the effects of the number of simulated unknown locations $n_m$ by varying it in $\{0, 10, 30, 70\}$. The results are given in Figure~\ref{fig_em}. First, the model trained with samples with $n_m = 10$ and $n_m = 30$ give lower MAEs compared to the model without random masking ($n_m = 0$). It indicates that randomly masking strategy can make the trained graph neural networks more generalizable to a completely new graph. Second, the model with a very large $n_m$ gives high MAEs, and the standard deviation of the evaluation errors during different training episodes are particularly high. This makes sense in that a very large $n_m$ will make the generated graph structures too fractured/sparse, makes it difficult for the graph neural networks to converge.

\section{Conclusion}
\label{C:5}
In this paper, we propose a novel spatiotemporal kriging framework---SATCN, which uses spatial graph neural networks to capture spatial dependencies and temporal convolutional networks to capture temporal dependencies. Specifically, SATCN features a masking strategy to forbid message passing from unobserved locations and multiple aggregators to allow the model to better characterize the spatial dependencies. We evaluate SATCN on diverse types of real-world datasets, ranging from traffic speed to global temperature. Our results show that SATCN offers superior performance compared with baseline models in most cases. SATCN is also robust against missing data, and it can still work well on datasets with missing data ratio up to nearly 50\%. SATCN is flexible in dealing with problem of diverse sizes in terms of the number of the nodes and the length of time window. This flexibility allows us to model time-varying systems, such as moving sensors or crowdsourcing systems. The masked spatial aggregation network proposed in this paper can also be viewed as a graph neural network for general missing data cases. This framework can be further integrated into time series forecasting frameworks with under missing data.

\ifCLASSOPTIONcaptionsoff
  \newpage
\fi

\bibliographystyle{IEEEtranN}%

\bibliography{acmart}

\begin{IEEEbiography}[{\includegraphics[width=1in,height=1.25in,clip,keepaspectratio]{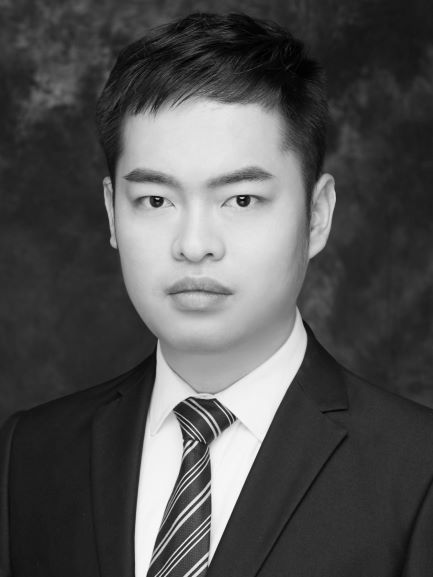}}]
{Yuankai Wu} received the PhD's degree from the School of Mechanical Engineering, Beijing Institute of Technology, Beijing, China, in 2019. He was a visit PhD student with Department of Civil \&
Environmental Engineering, University of Wisconsin-Madison from Nov. 2016 to Nov, 2017. He is a Postdoc researcher with Department of Civil Engineering at McGill University, supported by the Institute For Data Valorization (IVADO). His research interests include intelligent transportation systems, intelligent energy management and machine learning.
\end{IEEEbiography}

\begin{IEEEbiography}[{\includegraphics[width=1in,height=1.25in,clip,keepaspectratio]{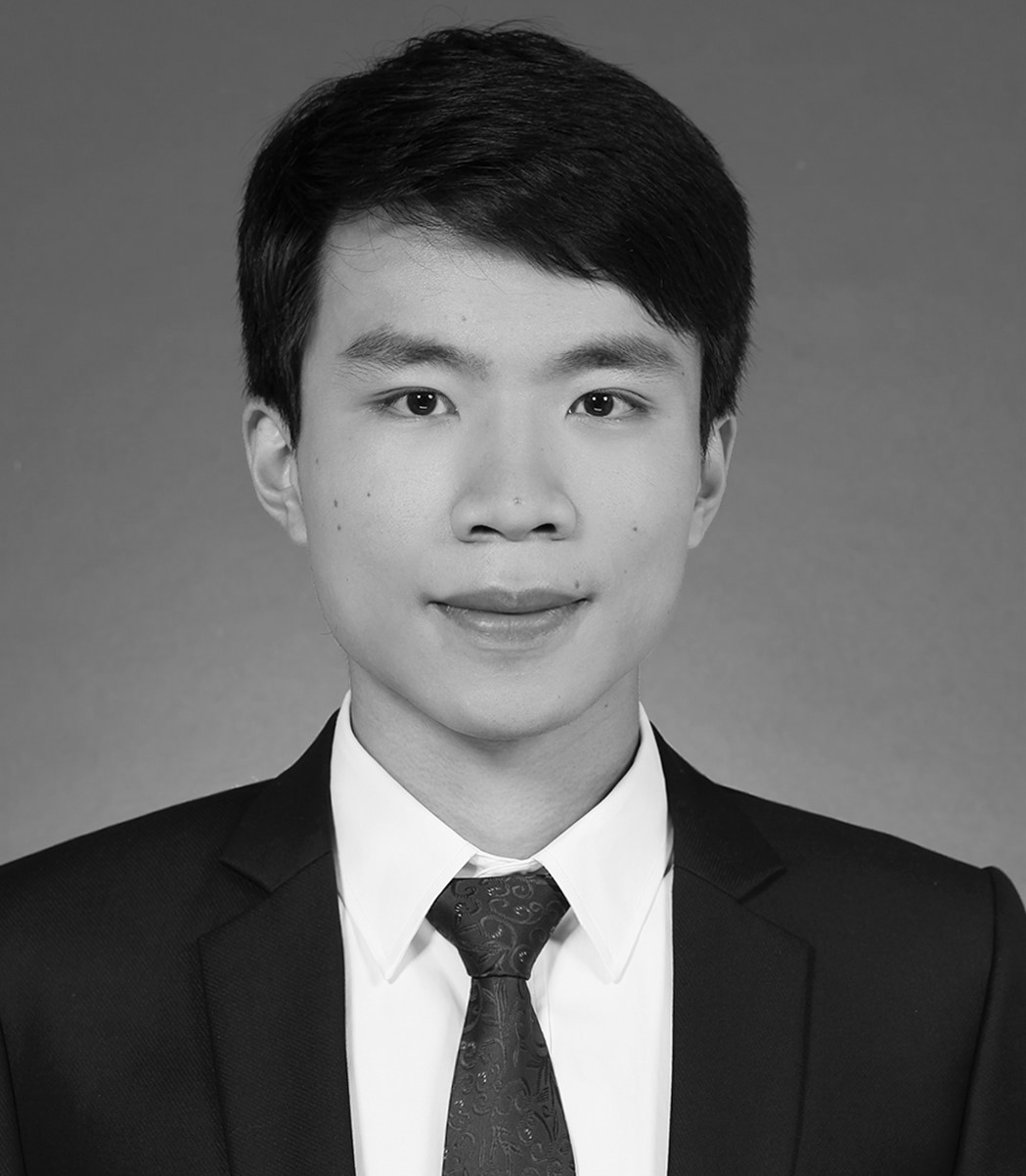}}]
{Dingyi Zhuang} received the B.S. degree in Mechanical Engineering from Shanghai Jiao Tong University, Shanghai, China, in 2019. He was a visit research assistant in the Department of Civil \& Environmental Engineering at National University of Singapore. He is a M.Eng. student in Transportation Engineering expected to graduate in 2021. His research interests lie in intelligent transportation systems, machine learning, and travel behavior modeling.
\end{IEEEbiography}

\begin{IEEEbiography}[{\includegraphics[width=1in,keepaspectratio]{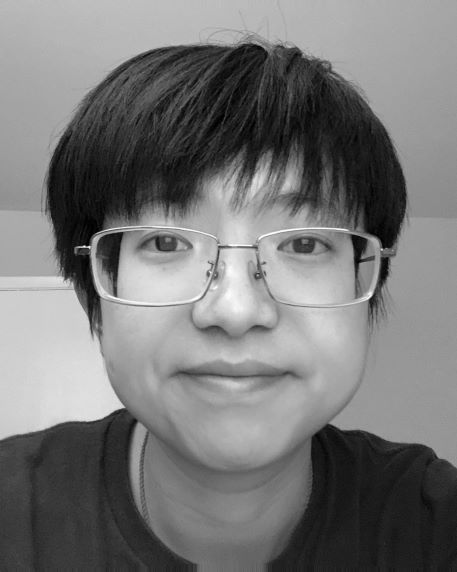}}]{Mengying Lei} received the B.S. degree in automation from Huazhong Agricultural University, in 2016, and the M.S. degree from the school of automation science and electrical engineering, Beihang University, Beijing, China, in 2019. She is now a Ph.D. student with the Department of Civil Engineering at McGill University, Montreal, Quebec, Canada. Her research currently focuses on spatiotemporal data modelling and intelligent transportation systems.
\end{IEEEbiography}

\begin{IEEEbiography}[{\includegraphics[width=1in,height=1.25in,clip,keepaspectratio]{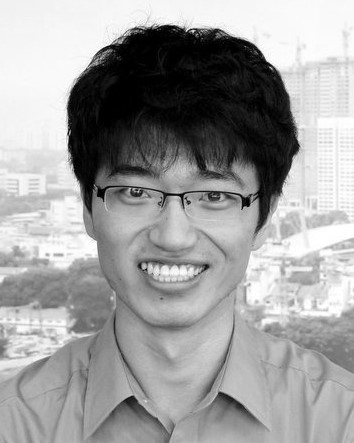}}]
{Lijun Sun} (Member, IEEE) received the B.S. degree in Civil Engineering from Tsinghua University, Beijing, China, in 2011, and Ph.D. degree in Civil Engineering (Transportation) from National University of Singapore in 2015. He is currently an Assistant Professor with the Department of Civil Engineering at McGill University, Montreal, Quebec, Canada. His research centers on intelligent transportation systems, machine learning, spatiotemporal modeling, travel behavior, and agent-based simulation.
\end{IEEEbiography}


\vfill

\end{document}